# LOCAL POST-HOC EXPLANATIONS FOR PREDICTIVE PROCESS MONITORING IN MANUFACTURING

Mehdiyev, Nijat, German Research Center for Artificial Intelligence (DFKI) and Saarland University, Saarbrücken, Germany, nijat.mehdiyev@dfki.de

Fettke, Peter, German Research Center for Artificial Intelligence (DFKI) and Saarland University, Saarbrücken, Germany, peter.fettke@dfki.de

## Abstract

*This study proposes an innovative explainable predictive quality analytics solution to facilitate the data-driven decision-making for process planning in manufacturing by combining process mining, machine learning, and explainable artificial intelligence (XAI) methods. For this purpose, after integrating the top-floor and shop-floor data obtained from various enterprise information systems, a deep learning model was applied to predict the process outcomes. Since this study aims to operationalize the delivered predictive insights by embedding them into decision-making processes, it is essential to generate the relevant explanations for domain experts. To this end, two complementary local post-hoc explanation approaches, Shapley values and Individual Conditional Expectation (ICE) plots are adopted, which are expected to enhance the decision-making capabilities by enabling experts to examine explanations from different perspectives. After assessing the predictive strength of the applied deep neural network with relevant binary classification evaluation measures, a discussion of the generated explanations is provided.*

# 1 Introduction

The recent proliferation of the internet of things (IoT), cyber-physical systems, cloud computing, enterprise information systems, and other innovative manufacturing-specific technologies, and consequently, the increasing availability of heterogeneous and voluminous production data, facilitate the manufacturing enterprises to establish data-driven intelligence (Lasi et al., 2014). Recognized as one of the key enablers of such manufacturing intelligence, machine learning, has been examined throughout the various stages of the manufacturing lifecycle, including design, evaluation, operation, and maintenance. These artificial intelligence based intelligent systems have already found extensive applications for different problems such as fault detection, predictive maintenance, operations planning, predictive energy consumption monitoring, predictive quality analytics, and decision support for various data-driven decision-making situations (Wang et al., 2018). In their survey-based investigation into large corporations, Brynjolfsson et al. (2011) have revealed that adopting data-driven decision-making results in a 5-6% increase in output and productivity. Furthermore, embedding data-driven decision-making in business processes yields a higher return on investment and equity, asset utilization, and market value (Provost and Fawcett, 2013).

A growing body of literature has further explicitly investigated the added value generated through successful adoption of another decisive technology, enterprise information systems, such as Enterprise Resource Planning (ERP), Customer Relationship Management (CRM), Supply Chain Management (SCM). The empirical findings reveal the necessity and utility of enterprise information systems for facilitating organizational data-driven decision-making (Hitt et al., 2002). To successfully deploy predictive analytics and establish a data-driven culture, manufacturing enterprises must ensure a consistent process and information flow and their horizontal and vertical integration along the entire value chain. Therefore, the seamless integration of the top-floor and shop-floor operations and data structures is a crucial point for data-driven intelligence, which constitutes a vital challenge. The manufacturing execution systems (MES) have been established as essential enterprise information systems that control, plan, and manage operational activities to address this issue. These operational level systems also provide an invaluable data source for manufacturing intelligence by capturing highly critical process details, including operational, material, machine, planning, personnel, energy, and quality data.

Advanced machine learning approaches developed on top of the data delivered by various information systems are the fundamental enablers of data-driven manufacturing intelligence and provide superior predictions and recommendations compared to intuitive forecasts and estimations. However, although there is overwhelming evidence to confirm the superiority of algorithmic reasoning, decision makers are often reluctant to use them, preferring instead to rely on the more vague human judgments (Fildes et al., 2009; Sanders and Manrodt, 2003). A number of studies have found that the decision makers' aversion to using predictions delivered by algorithms arises from intolerance to unavoidable errors. (Dietvorst et al., 2015). Therefore, such a reluctance can be overcome if these users can adjust when they suspect and understand why the algorithms err and fail (Dietvorst et al., 2018). Explainable Artificial Intelligence (XAI) has recently reemerged as a research domain that pursues the objective to address this issue and attempts to eliminate various other barriers to enable collaboration among AI-enabled intelligent systems and human users (Doshi-Velez and Kim, 2017). Novel XAI methods are currently being developed and used to deliver relevant transparency and justification mechanisms for establishing the required trust.

This study examines a process outcome prediction use-case from the individual manufacturing domain, particularly by focusing on an innovative predictive quality analytics problem. To put more precisely, this study attempts to develop an explainable process prediction solution by combining process mining, machine learning, and XAI approaches to enable data-driven decision-making. For this purpose, a black-box machine learning approach, deep neural network, is trained using various process and MES-specific features to predict the process outcomes, which is defined as the quality of the produced product for the given process sequence. Following this, two complementary local post-hoc ex-planation approaches, Shapley values, and ICE plots, are applied to generate the relevant explanations that facilitate the experts to justify the model decisions.

The remainder of this study is structured as follows: Section 2 introduces the background of explainable process prediction approaches by discussing the related work from AI, XAI, and Business Process Management. Section 3 introduces the used design science research method. Section 4 describes the methods used for designing the proposed artifact. The obtained results and explanations are presented in section 5. Section 6 discusses the practical and scientific implications and provides an overview of future work. Finally, section 7 concludes the study with a summary.

## 2 Related Work

### 2.1 AI and Business Process Management

Predictive process analytics, which is also referred to as predictive process monitoring or predictive business process management, has emerged as a promising branch of process mining to generate predictive insights by using the event log data delivered by process-aware information systems (PAIS) (Evermann et al., 2017; Di Francescomarino et al., 2018). Over time, an extensive literature has developed on predictive process monitoring by focusing on different prediction problems such as next event prediction, business process outcome prediction, remaining time prediction, prediction of activity delays, prediction of service level agreement violations, and other related problems (van der Aalst et al., 2011; Breuker et al., 2016; van Dongen et al., 2008; Evermann et al., 2017; Folino et al., 2012; Di Francescomarino et al., 2017, 2018; Lakshmanan et al., 2015; Le et al., 2014, 2017; De Leoni et al., 2016; Leontjeva et al., 2015; Márquez-Chamorro et al., 2017; Mehdiyev et al., 2017; Polato et al., 2014, 2018; Tax et al., 2017; Unuvar et al., 2016; Wynn et al., 2014).

However, a thorough analysis of these studies reveals that they mainly use the process data generated from the management-level enterprise information systems such as ERP, CRM, or SCM. Only a few studies have examined the applicability of process mining techniques such as process discovery, conformance checking, predictive process analytics, or process enhancement in the manufacturing domain by concentrating on the event log data delivered by MES (Fettke et al., 2020; Gröger et al., 2012). This study aims to fill this research gap by proposing a relevant explainable predictive process analytics solution. Furthermore, the literature pertaining to predictive process monitoring suggests that the black-box machine learning approaches provide superior predictive performance compared to traditional comprehensive methods (Di Francescomarino et al., 2018).

A considerable amount of research has recently focused on applying deep feedforward neural networks, convolutional neural networks (CNN), long-short term memory networks (LSTM), generative adversarial nets, and other deep learning architectures for predictive process analytics and monitoring (Evermann et al., 2017; Al-Jebrni, Cai and Jiang, 2018; Mehdiyev, Evermann and Fettke, 2018; Tello-Leal, Roa, Rubiolo and Ramirez-Alcocer, 2018; Camargo, Dumas and González-Rojas, 2019; Di Mauro, Appice and Basile, 2019; Theis and Darabi, 2019; Weinzierl, Wolf, et al., 2020; Kratsch, Manderscheid, Röglinger and Seyfried, 2020; Park and Song, 2020; Taymouri et al., 2020, Neu et al., 2021).

### 2.2 Explainable Artificial Intelligence (XAI)

Although deep learning approaches deliver more precise prediction outcomes, their lack of explanation constitutes practical challenges for establishing data-driven decision-making (Guidotti et al., 2018). XAI has recently gained increasing attention as a research discipline to make the communication between intelligent systems and human users understandable and establish trust in non-transparent models or their outcomes (Doshi-Velez and Kim, 2017; Gunning and Aha, 2019; Miller, 2019). Various studies provide valuable insights into different aspects of XAI research questions, for instance, by providing an overview of possible taxonomies for explanation techniques (Gilpin et al., 2018; Guidotti et al., 2018; Lipton, 2018), presenting different mechanisms and approaches for evaluating the expla-nations (Doshi-Velez and Kim, 2017), discussing the objectives of XAI solutions (Nunes and Jannach, 2017), defining the stakeholders of the explanation methods (Preece et al., 2018), introducing the relevant insights from social sciences (Miller, 2019), and proposing the necessity of considering the findings from cognitive sciences (Fürnkranz et al., 2019).

Recently, various studies have been conducted on making predictive process analytics explainable by using various local and global post-hoc explanation approaches (Rehse, Mehdiyev and Fettke, 2019; Mehdiyev and Fettke, 2020a, 2020b; Rizzi, Di Francescomarino and Maggi, 2020; Weinzierl, Zilker, et al., 2020).

## 3    Research Method

To design an explainable process outcome prediction solution, we follow the design science research (DSR) approach proposed by Peffers et al. (2007) which combines principles, practices and procedures to conduct applied research. Widely used in information systems research, the adopted DSR comprises six essential steps: problem identification, defining objectives of a solution, design and development, demonstration, evaluation, and communication of the novelty and rigor to the relevant audience. The description of the details of each step is as follows:

- **Problem Identification and Motivation:** A robust and reliable process planning solution should support the domain experts in defining the sequence of operations and processes for producing the construction parts, identifying the necessary and relevant resources, conducting time, cost, and risk-specific estimations, defining preventive measures proactively monitoring the deviations from the desired outcomes. The ability to predict the quality of the produced products using the process-specific features is another critical capability that generates proactive recommendations for adapting and optimizing the operational processes.

    The integration of various top-floor PAIS with shop-floor systems over various stages of the production automation pyramid enables carrying out such process analytics activities by recording and delivering the relevant data. Subsequently, AI-enabled predictive process analytics methods should be concepted, developed, and implemented to realize the concept of data-driven process analytics in manufacturing. Nevertheless, the most precise, reliable predictive process monitoring methods are non-linear black-box approaches that fail to generate the relevant explanations about their outcomes or inferencing process due to their non-transparent nature. This, in turn reduces the trust in the underlying artificial advice-givers and introduces the acceptance barriers in using data-driven process-specific solutions even though they often deliver more superior outcomes than intuition-based decisions.

- **Objective of a Solution:** The proposed solution aims to embed data-driven decision-making into the predictive process planning by formulating the predictive quality analytics as a process outcome prediction problem. To generate a trustworthy, consistent, and sufficient predictive process analytics solution for a manufacturing firm, the analysts should ensure that high-quality data from relevant enterprise information systems are acquired, integrated, aggregated, and pre-processed. Consequently, sound machine learning models with plausible predictive strength should be constructed and validated.

    Another important objective of the proposed solution is to operationalize the recommendations and insights delivered by adopted advanced machine learning models for process predictions by making them explainable and interpretable. Generating explanations for black-box machine learning approaches pose challenges since various properties of the decision-making environment, the target audience's requirements, and other economic, organizational, and legal considerations influence the appropriateness of explanations significantly. Therefore, to systematically conduct the explanation generation process, we follow the conceptual framework proposed by Mehdiyev and Fettke (2020b), which was designed to guide developing explainable process prediction solutions.

According to this framework, it is crucial to identify the target audience of the explanations, define their objectives, examine the context of the explanation situation, and choose the relevant techniques to facilitate the users to attain their goals. In this study, the domain experts with deep expertise in process and production planning but with limited machine learning experience are the primary target audience. These users prefer to justify the outcomes delivered by non-linear machine learning models rather than understand the complicated inner working mechanism of these opaque models. Such a ratification goal of the model recommendations enables the experts to verify whether the model findings conform to their knowledge and expertise, learn complex relationships, especially in weak theory domains, and identify the preventive measures. To this end, it is conceivable to suggest that post-hoc explanation tools are the most appropriate tools which attempt to open the black-box models once the models are already trained. Since the domain experts in our case are more interested in understanding every single model decision, we propose adopting the local post-hoc explanation approaches.

- **Design and Development:** The proposed design artifact, explainable process outcome prediction solution, consists of the deep feedforward neural network trained using process-specific features delivered by the relevant information systems and two local-post explanation techniques, Shapley values and ICE plots. These explanation techniques are chosen since the target audience was defined as domain experts interested in justification of individual model outcomes. Training and validation of the applied deep learning approach were carried out using the open-source "H2O" machine learning platform. For generating both local post-hoc explanations, the "iml: Interpretable Machine Learning" package was used. Data visualization was performed using packages such as ggplot2 and ROCR were used. All modeling and experimentation activities were conducted using the R programming language.

- **Demonstration:** This study examines a real-world use-case study with semi-artificial data obtained from job production processes to demonstrate the applicability of the proposed explainable process prediction solution.

- **Evaluation:** To evaluate the performance of the deep learning classifier, we calculated various threshold-free evaluation measures and single-threshold binary classification evaluation measures. A discussion on the evaluation of the explanation solutions was provided as well.

- **Communication**: The communication of the proposed solution and its outcomes are performed through publications, prototype demonstrations and presentations for different audiences such as scholars, production experts, and data engineers/analysts.

# 4 Local Post-Hoc Explanations for Process Outcome Prediction

This section provides an overview of the steps carried out to develop an explainable predictive process analytics solution. The initial step for developing the proposed manufacturing intelligence system is extracting machine learning features after aggregating, cleaning, transforming, normalizing, and harmonizing the data obtained from the relevant enterprise information systems used for managing operational processes. In particular, the control-flow, data-flow and time-specific process features are generated from the event log data supplied by MES. After carrying out comprehensive data preprocessing and feature engineering procedures, black-box machine learning methods are applied to generate the relevant process outcome predictions. The process outcome prediction problem is formulated as predicting the class label of interest for a particular (running) process sequence by learning from the historical cases with the relevant labels. In this study, the quality inspection results for the production parts produced as an outcome of a sequence of particular processes are used to define the relevant process outcome labels (Passed vs. Failed). After having verified and validated the predictive performance of the adopted deep learning approach, two local post-hoc explanation approaches, Shapley values and ICE, are used to explain every single outcome of the model for domain experts.

## 4.1 Use-Case and Data Preparation

To demonstrate the applicability of the proposed explainable process prediction solution, we investigate the process outcome prediction use-case in a medium-sized manufacturing firm operating in the field of tool and fixture construction. Our proposed approach is a final component of a more complex process planning and analytics solution developed within a consortium research project. This study pursues the aim to use MES-driven process-specific features such as the total number of process steps required to produce the planned manufacturing part, average duration per process step, average energy consumption per process step, planned setup time, Overall Equipment Effectiveness (OEE), employee effectiveness, etc. as input variables to predict whether the quality of the produced parts fulfill the pre-defined requirements or not. Due to the limited availability of actual production data at this stage of the research project, we use semi-artificially generated data partially based on the initial input delivered by the implemented MES at the partner manufacturing firm and feedback from process experts to achieve as close as possible data structures reflecting the actual situation of examined processes. We adopt the data generation approach based on radial basis function networks proposed by Robnik-Šikonja (2015), which learns sets of the Gaussian kernels and uses them to generate data from the same distributions. The discussion of the following steps, training advanced but opaque machine learning models, and generating the explanations for the model outcomes using the post-hoc explanation techniques are introduced in the following subsections.

## 4.2 Deep Feedforward Neural Networks for Process Outcome Prediction

As has been previously suggested in the literature on the explanations for intelligent systems, an inaccurate explanation can be misleading and is worse than no explanation at all (Swartout and Moore, 1993). In this context, Zhao and Hastie (2019) suggest that the first prerequisite for successful explanations is the solid predictive model. In this study, we aim to address the process outcome prediction measured as the quality of the produced construction parts using the information extracted from the relevant process sequences. Considering its superiority over conventional techniques when working with the tabular data, we adopted a deep feedforward neural network for the underlying problem. (Candel et al., 2016). To avoid overfitting problem, we used early stopping and dropout techniques. A summary of these regularization techniques and an overview of the adopted deep learning approach parameters are presented in Table 1.

| Parameter | Value |
|---|---|
| Initial Weight Distribution | Uniform Adaptive |
| Activation Function | Rectifier with Dropout |
| Input Dropout Ratio | 0.2 |
| Hidden Layer Dropout Ratio | 0.5,0.5,0.5,0.5 |
| Epochs | 1000 |
| Adaptive Learning Rate Algorithm | ADADELTA |
| Rho (adaptive learning rate time decay factor) | 0.99 |
| Epsilon (adaptive learning rate time smoothing factor) | 1e-8 |
| Max w2 (the constraint for the squared sum of the incoming weights per unit) | 100 |
| Early Stopping Metric | AUROC |
| Stopping Rounds | 5 |
| Stopping Tolerance | 0.005 |

*Table 1.       The Parameters of Deep Feedforward Neural Networks*

### 4.3 Shapley Values

As one of the widely recognized feature-attribution-based explanation techniques, the goal of Shapley values is adapting the cooperative game theory to machine learning interpretation. In this context, the values of the input features for the examined instance are treated as game players, and the classification model prediction scores as the corresponding payoffs. The main idea behind this additive feature attribution method is measuring the relevance and importance of a particular examined feature by estimating the accuracy of every combination of all other features in the absence of the examined feature and then measuring how adding the examined feature to each combination increases the accuracy. The interpretation of the Shapley values is intuitive, which can be formulated as follows: Shapley value is the contribution of the given feature value to the difference between the actual prediction for the examined instance and the mean prediction for the dataset. In other words, according to the Shapley value-based explanation, the difference between the actual prediction for the examined instance an average prediction is perfectly distributed among all features (Aas et al., 2019). One of the essential superiorities of the Shapley values compared to alternative perturbation-based feature attribution techniques is its axiomatic uniqueness (Bhatt, Xiang, et al., 2020). This approach is one of the few explanation methods which has a strong theory behind it.

The most critical challenge for operationalizing the explanations with Shapley values is that the computational complexity for calculating the exact values grows exponentially when the underlying machine learning problem has many features (Aas et al., 2019). Various approximation methods have been proposed to address this challenge, using different approaches such as Monte Carlo approximation, weighted linear regression, etc. (Lundberg et al., 2018; Lundberg and Lee, 2017; Štrumbelj and Kononenko, 2014).

### 4.4 Individual Conditional Expectation (ICE) Plots

The Shapley values method provides beneficial information on the contribution of each feature value to the model outcome; however, it has a static nature that provides only the snapshot of the situation by examining the given feature values. By introducing the ICE plots, we can generate complementary explanations that allow us to examine how the predictions of each observation change when the feature values change (Goldstein et al., 2015). Combining these two local post-hoc explanation techniques is presumed to provide a more comprehensive explanation for justifying the model outcomes.

The main working principle of ICE plots is also intuitive. For the examined instance, we choose a plot variable of interest every time and generate new instances by changing its value by using the values from the pre-defined grid and keeping all other variables constant. The generated instances are then fed to the underlying black-box model, and obtained prediction scores over different plot feature values are visualized. ICE plots can be seen as the local version of the popular global post-hoc explanation approach, Partial Dependence Plots (PDP), which shows the marginal effect of the chosen variables on outcome delivered by the underlying machine learning approach (Friedman, 2001). According to the recent findings by Zhao and Hastie (2019), this approach can be used to generate causal explanations for black-box models.

## 5 Results

### 5.1 Experiment Settings

To assess the predictive strength of the adopted deep neural networks, we compute and introduce threshold-free evaluation measures such as area under the Receiver Operating Characteristic Curve (AUROC) and area under the Precision-Recall Curve (AUPRC) and various single-threshold measures. The relevant information from the obtained confusion matrix, which summarizes the classifier's performance, is used to compute the single-threshold binary classification measures. The main categories of such a contingency table in our use-case are (i) true positives *(tp)* which are the products with good quality (Passed) that were detected correctly by the algorithm, (ii) true negatives *(tn),* which refers to the ability of the classifier to identify those products with the bad quality correctly (Failed), (iii) false positives *(fp)* are the products with bad quality but classified as good products and (iv) false negatives *(fn)* are the products with bad quality but identified by the model as having good qualities.

## 5.2 Performance of Black-Box Model

The obtained AUROC (0.965) and AUPRC (0.967) values suggest that the applied machine learning technique achieves reliable predictive performance and can detect the quality of the produced parts effectively (see Figure 1).

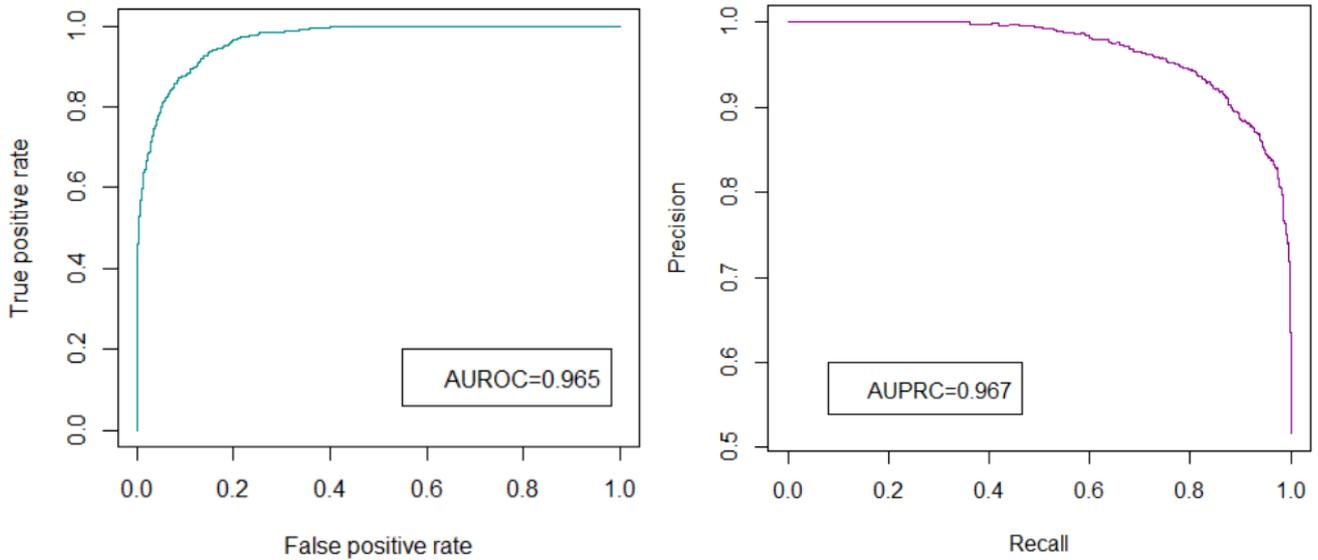

*Figure 1.*     Areas Under ROC and PRC Curves Obtained by the Deep Learning Model

Furthermore, various single-threshold binary classification evaluation measures obtained at the defined cut-off threshold of 0.409 at which Matthews correlation coefficient (MCC) is maximized confirm the adopted deep neural network's predictive strength for the underlying problem (see Table 2).

| Evaluation Measure | Formula | Value |
|---|---|---|
| F1-Measure | $\frac{2tp}{(2tp + fp + fn)}$ | 0.901 |
| Accuracy | $\frac{tp + tn}{tp + tn + fp + tp}$ | 0.894 |
| Precision | $\frac{tp}{tp + fp}$ | 0.869 |
| Recall | $\frac{tp}{tp + fn}$ | 0.935 |
| Specificity | $\frac{tn}{(tn + fp)}$ | 0.849 |
| Absolute Matthews correlation coefficient (MCC) | $\frac{tp * tn - fp * fn}{\sqrt{(tp + fp) * (tp + fn) * (tn + fp) * (tn + fn)}}$ | 0.789 |
| False Negative Rate | $\frac{fn}{fn + tp}$ | 0.064 |
| False Positive Rate | $\frac{fp}{tn + fp}$ | 0.150 |

*Table 2.*     Binary Classification Evaluation Measures and Obtained Values

## 5.3 Local Post-Hoc Explanations

After verifying and validating the performance of the adopted neural networks approach, we can now generate the relevant local post-hoc explanations for process domain experts and production planners. This subsection introduces Shapley values and ICE plots for randomly chosen observations with true negative (Failed the quality inspection and identified correctly by the model) and true positive (Passed the quality inspection and identified correctly by model) predictions. Furthermore, a discussion on harmonic transitions between chosen local post-hoc explanation approaches is provided, which is presumed to enhance experts' decision-making capabilities by enabling them to examine the explainable process predictions from different perspectives.

### 5.3.1 Shapley Values

Figure 2 introduces the Shapley values for an observation from the dataset with a true negative prediction, which implies that the deep learning model successfully predicted that the product quality is not at the desired level. With a prediction probability of passing the quality test of 0.14, the prediction score for the examined instance significantly below the average prediction of 0.47. The respective Shapley value reveals that the Overall Equipment Effectiveness (OEE) (0.44) in the examined instance decreases the probability for the "Passed" significantly by 27%. The observed value of another MES-specific variable, the employee productivity (0.47), pushes the probabilities towards a negative direction by increasing the probability for class "Failed". The given values of the planned setup time and the total number of the process steps influence the probabilities of passing quality tests slightly negatively. Although all other variables, including planned production duration, average energy consumption per process step, and average duration per process step increase the probability in favor of the positive class, passing the quality test, their contribution are too small to change the model's decision for the examined instance.

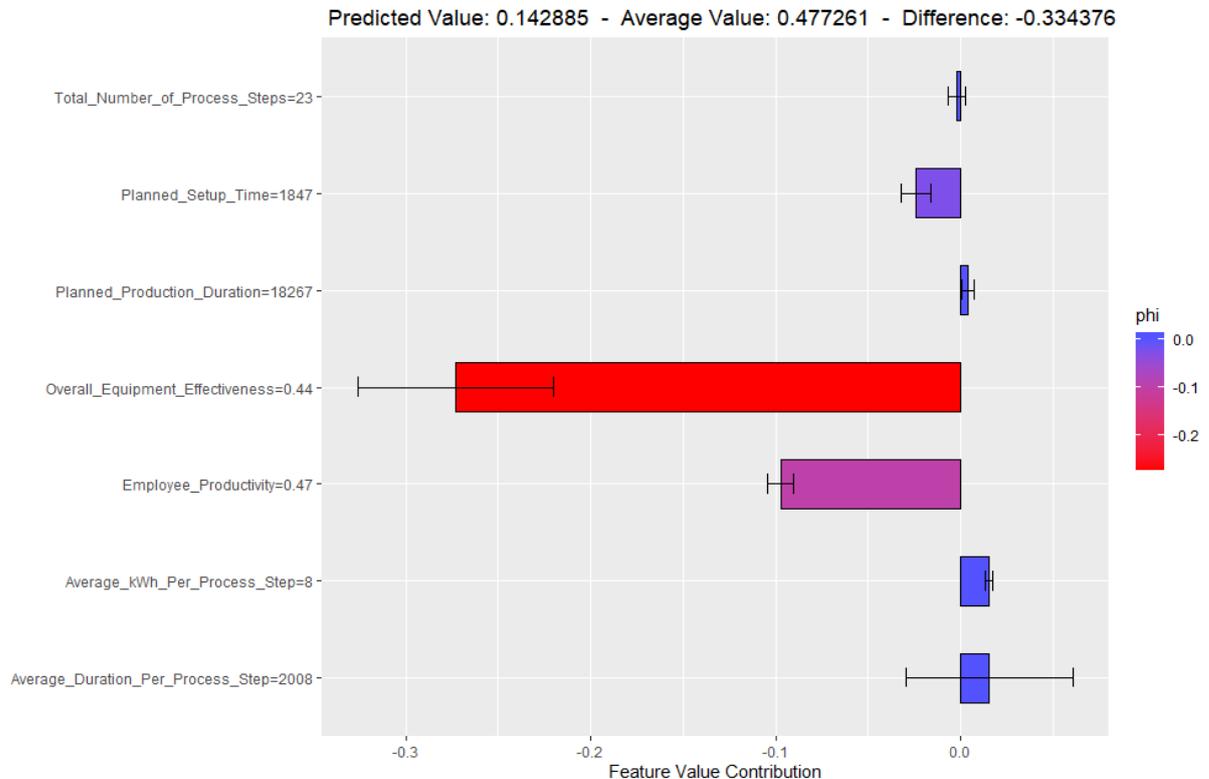

*Figure 2.     Shapley Values for the Observation with the True Negative Prediction*

Figure 3 introduces the Shapley values obtained for the observation, which was correctly predicted as "Passed" with a prediction score of 0.84. The sum of the recorded parameter values' contributions yields the difference between actual and average prediction with almost 0.37. According to the obtained Shapley values, the high value of OEE in the examined instance (0.95) is strongly associated with a high prediction score in favor of class "Passed". This value increases the probabilities of passing the quality test by 32%. Furthermore, the higher value of employee productivity (0.82) has also a positive impact on the prediction scores. Moreover, the given values of planned setup time (1767) and average duration per process step (2220) increase the probability of passing the test. In this examined instance, the total number of process steps, the planned production duration, and the average energy consumption per process step decrease the probabilities for passing the quality test but are not high enough to change the model's correct decision.

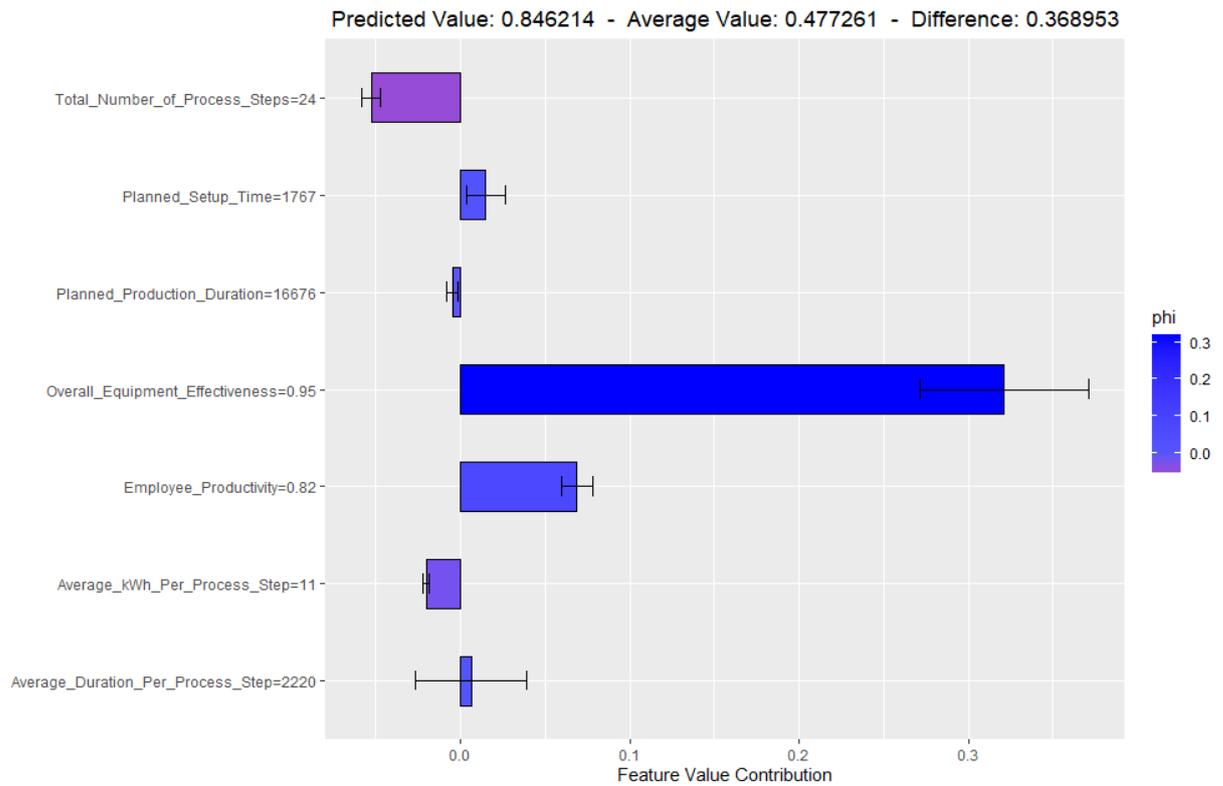

*Figure 3.* Shapley Values for the Observation with the True Positive Prediction

### 5.3.2 Individual Conditional Expectation (ICE) Plots

For illustrative purposes, the ICE curves for chosen plot variables, OEE, employee productivity, and average duration per process step are presented in this subsection. Figure 4 suggests that in both examined instances, an increase in the OEE values has a positive impact on prediction scores for good product quality (Passed). This finding aligns with the Shapley-based explanation introduced above. The green ICE curve, which represents the observation with true positive prediction, implies that the OEE value of 0.95 results in a high prediction score around 0.84, as also described above in the Shapley Value explanations (see Figure 3). This feature is so crucial for this observation that a sharp decrease in its values may decrease the probabilities and lead to a negative outcome, failing the quality test. For instance, keeping all other variables constant, using a production machine with an OEE value of 0.4 instead of the current machine with an OEE of 0.95 in the examined instance would have resulted in bad product quality. Simultaneously, the ICE curve analysis for the observation with true negative prediction (presented with red line) suggests that a considerable increase in its current value (0.44) may switch the model decision to a positive outcome.

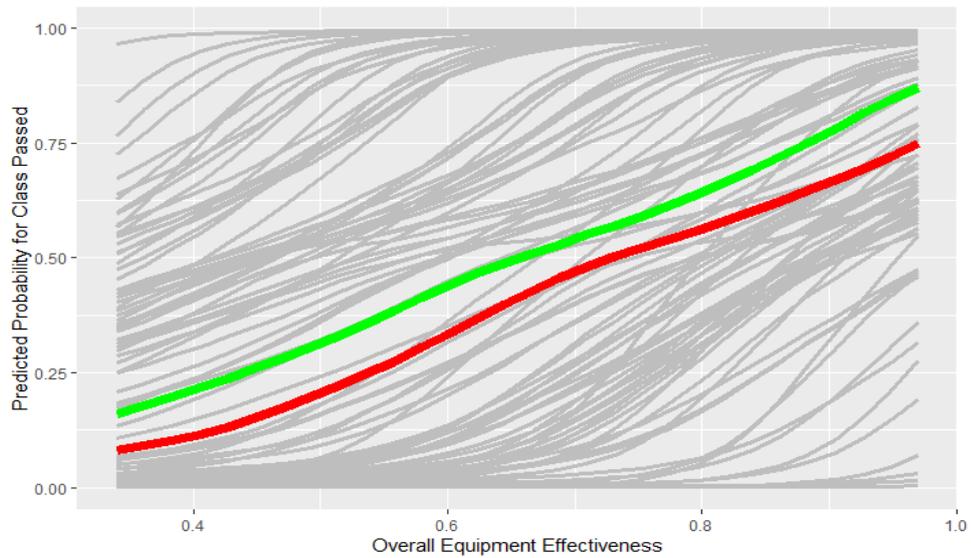

*Figure 4.        ICE Plots for Overall Equipment Effectiveness*

As depicted in Figure 5, the similar trend can be observed in both ICE plots for the employee production feature. An increase in the values of these variables increases the prediction probability of being classified as "Passed", positive class. This interpretation can also be easily linked to the Shapley explanations. This feature's value for the observation with the true negative prediction (0.47) is considerably lower than the feature value in the other examined instance (0.82). Therefore, in the first case, it has negative Shapley values, whereas the Shapley value in the second case pushes the probabilities in favor of the positive class.

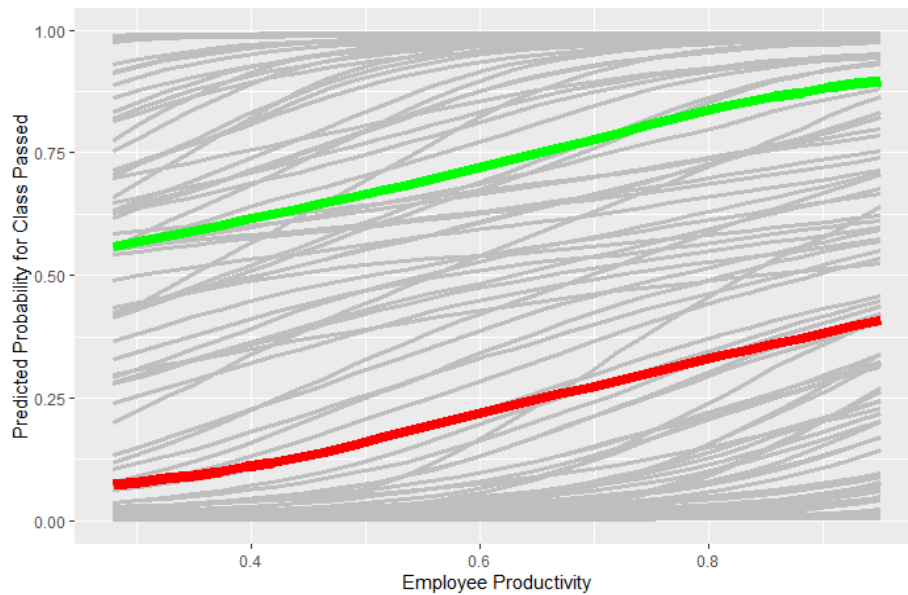

*Figure 5.        ICE Plots for Employee Productivity*

Finally, Figure 6 presents the ICE plots for the average duration per process step. Both ICE plots (red and green lines) follow a similar trend. In contrast to the ICE plots for two previously examined features, an increase in the average duration per process step decreases the prediction scores for good quality. This feature's values are close for both observations (2220 vs. 2008 seconds) and are positioned at a critical point because an increase after around 2000 seconds results in a sharper decrease of the prediction scores for class "Passed". By using the explanations generated by these ICE plots, the domain experts can adapt the process plans by defining relevant strategic measures, which may lead to outcomes with higher qualities.

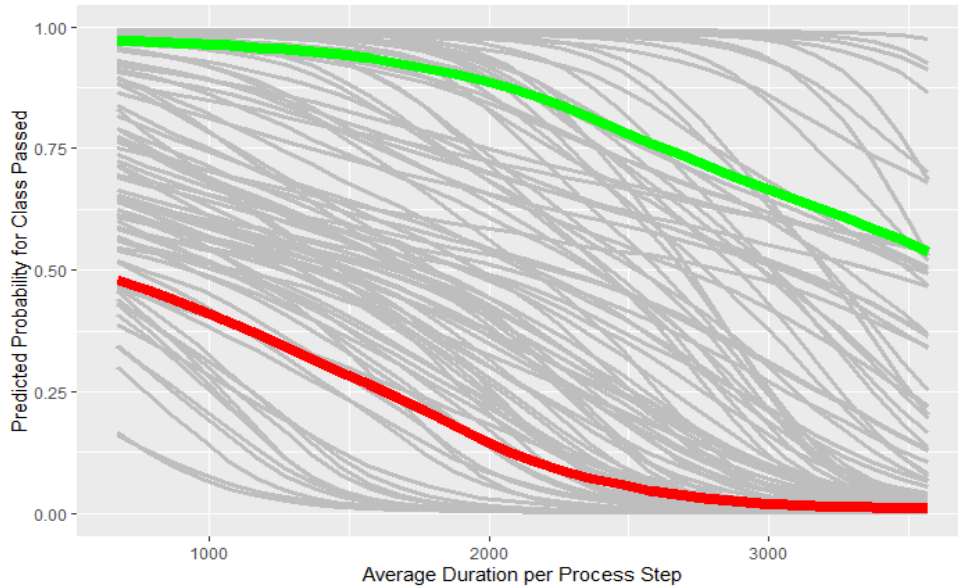

*Figure 6.      ICE Plots for Average Duration per Process Step*

# 6      Discussion

The innovation contribution of this study lies in introducing an emerging and innovative AI use-case in the manufacturing domain by proposing an operational process data-driven analytics system with explanation components which has been very superficially examined in prior studies. Combining the advanced black-box machine learning method, process mining-based feature engineering, and widely recognized local post-hoc explanation techniques could produce a reliable and plausible solution for predictive process analytics. To this end, after analyzing the explanation situation carefully and examining the profiles and expectations of the system users diligently, we have demonstrated the appropriateness of two complementary local post-hoc XAI approaches, Shapley values, and ICE plots. These local explanations are sought to facilitate experienced decision-makers to justify the model outcomes, understand unexpected results, and resolve anomalies. Furthermore, their cognitively simplistic representations also enable novice users to learn from the system or clarify causal relationships without placing high cognitive loads on them.

It is also interesting to note that with relevant adjustments, it is possible to obtain global explanations from both local post-hoc explanation techniques, which are relevant for more strategic decisions such as enhancing processes or making investment decisions in production lines or facilities. Shapley values can be aggregated and deliver various global post-hoc explanations such as SHAP dependence plots, SHAP summary plots, and SHAP-based feature importance. The global version of ICE plots is the PDPs which was discussed above. In this study, we have investigated the possibilities of generating explanations for the applied deep neural network for a classification task. However, it is worth mentioning that both applied local post-hoc explanations and their global extensions are model agnostic approaches. This fact implies that they can be easily used for explaining other black-box approaches such as tree ensembles or other deep learning architectures. Moreover, these techniques support the explanation generation process for regression problems.

Our study verified the predictive strength of the applied deep feedforward neural network in terms of different threshold-free and single threshold classification evaluation measures. However, it is also vital to examine and assess the appropriateness of generated explanations. According to Doshi-Velez and Kim (2017), human grounded evaluation, application grounded evaluation, and functionality grounded evaluation are alternative approaches for evaluating the generated explanations. A few attempts have recently been made aimed at evaluating the Shapley explanations. One of the initial human-grounded evaluations of the Shapley values was made by Weerts (2019), who couldn't find out whether Shapley explanations significantly impact task utility measured as task effectiveness and mental efficiency. Bhatt, Weller et al. (2020) have recently conducted research on evaluating the explanation approaches in terms of three desiderata, namely low sensitivity, high faithfulness, and low complexity. According to their results, Shapley values are the most faithful explanation function for small datasets and the least sensitive approach compared to different explanation techniques such as DeepLift, Gradient Saliency, Integrated Gradients etc. To further our research, we plan to refine and develop the relevant evaluation mechanisms and perform the assessments of the suitability of the proposed explainable process prediction solution in the production facilities with real users in terms of various desiderata.

It is worth mentioning that in the scope of the underlying research project, we also examine the applicability of various alternative local post-hoc explanation approaches such as LIME, counterfactual explanations, case-based explanations and investigate their harmonic combination with applied techniques in this study. Furthermore, our future research work also pursues the objective of developing novel local post-hoc explanation approaches to overcome the various shortcomings of perturbation-based post-hoc explanation techniques.

## 7     Conclusion

This study's main objective was to propose an explainable process prediction solution to facilitate data-driven decision-making for process analytics in manufacturing. After a thorough data preparation phase, a deep neural network approach was applied to predict the process outcomes according to the defined quality criteria. The obtained AUROC (0.965) and AUPRC (0.967) and relevant single-threshold measures, particularly the absolute MCC (0.789) and F1-Measure (0.901), suggest that the applied machine learning technique achieves strong predictive performance that fulfills the predefined success criteria defined by process owners. After verifying the model performance and consequently defining the target audience and their justification purposes, we generated the relevant explanation using two complementary local post-hoc explanation approaches, Shapley values and ICE plots. Their harmonic use allows the decision-makers to examine the model outcomes from different perspectives. This study is one of the initial attempts to make production process data-driven predictions explainable, and further efforts for developing alternative explanation solutions are required to enhance the acceptability of AI in manufacturing.

**Acknowledgment:** This research was funded in part by the German Federal Ministry of Education and Research under grant number 01IS19082A (project KOSMOX).